\documentclass[10pt, conference, letterpaper]{IEEEtran}
\IEEEoverridecommandlockouts
\usepackage{cite}
\usepackage{amsthm}
\usepackage{amsmath,amssymb,amsfonts}
\usepackage[cmintegrals]{newtxmath}
\usepackage{algorithm}
\usepackage{algorithmicx}
\usepackage{enumerate}
\usepackage{algpseudocode}
\usepackage{booktabs}   
\usepackage{multirow}    
\usepackage{graphicx}   
\usepackage{caption}
\usepackage{bm}
\usepackage{subfig}
\usepackage{graphicx}
\usepackage{array}
\usepackage{textcomp}
\usepackage{xcolor}
\usepackage{url}
\usepackage{tabularx}
\usepackage{color}
\makeatletter
\usepackage{enumitem}
\setlist{nosep} 
\usepackage[font=small, skip=3pt]{caption}
\makeatother
\newlength{\whilewidth}
\algdef{SE}[parWHILE]{parWhile}{EndparWhile}[1]
  {\parbox[t]{\dimexpr\linewidth-\algmargin}{
     \hangindent\whilewidth\strut\algorithmicwhile\ #1\ \algorithmicdo\strut}}{\algorithmicend\ \algorithmicwhile}%
\algnewcommand{\parState}[1]{\State
  \parbox[t]{\dimexpr\linewidth-\algmargin}{\strut #1\strut}}

\begin{document}
\title{Inductive Spatio-Temporal Kriging with Physics-Guided Increment Training Strategy for Air Quality Inference}    

 \author{Songlin Yang,
Tao~Yang,~\IEEEmembership{Member,~IEEE,}
and~Bo~Hu,~\IEEEmembership{Member,~IEEE}      
}
\maketitle

\begin{abstract}
The deployment of sensors for air quality monitoring is constrained by high costs, leading to inadequate network coverage and data deficits in some areas. Utilizing existing observations, spatio-temporal kriging is a method for estimating air quality at unobserved locations during a specific period. Inductive spatio-temporal kriging with increment training strategy has demonstrated its effectiveness using virtual nodes to simulate unobserved nodes. However, a disparity between virtual and real nodes persists, complicating the application of learning patterns derived from virtual nodes to actual unobserved ones. To address these limitations, this paper presents a Physics-Guided Increment Training Strategy (PGITS). Specifically, we design a dynamic graph generation module to incorporate the advection and diffusion processes of airborne particles as physical knowledge into the graph structure, dynamically adjusting the adjacency matrix to reflect physical interactions between nodes. By using physics principles as a bridge between virtual and real nodes, this strategy ensures the features of virtual nodes and their pseudo labels are closer to actual nodes. Consequently, the learned patterns of virtual nodes can be applied to actual unobserved nodes for effective kriging.
\end{abstract}

\begin{IEEEkeywords}
Air quality inference, sensors, inductive spatio-temporal kriging, physics principles, increment training strategy
\end{IEEEkeywords}

\section{Introduction}
As industrialization advances and urbanization accelerates, air pollution has become a critical issue affecting public health \cite{feng2024air}. Detailed spatial information on urban air quality is essential for the residents. Currently, air quality data collection primarily relies on monitoring stations managed by the government. These stations are outfitted with Internet of Things (IoT) sensors, enabling precise measurement of air quality indicators at targeted locations \cite{zaidan2020intelligent}. However, the widespread deployment of these sensors faces enormous challenges, particularly high costs, which result in insufficient network coverage and data gaps in certain areas. These limitations hinder the public from identifying potential environmental risks. Therefore, utilizing observed data to estimate air quality in areas without monitoring stations is crucial.

The fine-grained inference of air quality is principally addressed through two methods: physical and data-driven approaches. Based on theoretical assumptions, physical methods estimate air quality distribution by simulating the propagation of pollutants in the atmosphere \cite{yang2019new, kim2012urban}. Nevertheless, these idealized assumptions might diverge from real-world conditions, leading to prediction discrepancies. As an appealing alternative, data-driven methods estimate air quality by analyzing multisource urban data and capturing the spatio-temporal correlations of air quality distribution. Early statistical methods address this issue with multiple linear regression \cite{hoek2008review} and Gaussian Process regression \cite{li2011review}. Although these models are easy to implement, they cannot handle complex nonlinear dependencies. In recent years, deep learning-based approaches have garnered substantial attention due to their enhanced expressive capabilities \cite{cheng2018neural, han2021fine, hu2023graph}. For instance, some studies adopt deep neural networks (DNN) to integrate external data (e.g., Points of Interest (POI), road networks, and meteorological data) and dynamically adjust the contributions of different monitoring stations to unobserved areas through attention mechanisms \cite{cheng2018neural, han2021fine}. Nonetheless, these models rely on scenario-specific data that is not readily available, which limits their generalization capabilities in dynamic scenarios. As a robust substitute, inductive spatio-temporal kriging has recently gained prominence in air quality inference \cite{wu2021inductive, hu2023decoupling, zheng2023increase}. This method leverages graph neural networks (GNN) to manage complex spatio-temporal relationships and employs inductive learning strategies to reduce reliance on external data. The inductive learning mechanism enables the model to learn general patterns from training data and apply this knowledge to new, unseen scenarios \cite{wu2021inductive}. Consequently, these models can perform kriging for various unobserved locations without retraining.

The inductive model primarily employs the following training settings: (i) constructing a graph structure on the observed nodes, (ii) randomly masking some nodes to simulate unobserved nodes that require inference, and (iii) training the model using the unmasked nodes to reconstruct the values of the masked nodes, thereby equipping the model with inference capabilities for unobserved nodes. Unfortunately, in this context, the graph used for training is sparser than the graph used for inference, as the inference graph includes additional unobserved nodes requiring inference, leading to the so-called graph gap issue. KITS \cite{xu2023kits} proposes an increment training strategy to solve this problem. This strategy introduces virtual nodes into the training graph, utilizing these nodes to simulate the behavior and characteristics of unobserved nodes, thereby ensuring equal nodes in the training and inference graph. 

Although the increment training strategy mitigates the graph gap issue, it faces fitting problems due to the disparity between virtual and real nodes. These virtual nodes are not directly mapped to real-world physical entity sensors and lack the perception of the environment changes. In contrast to real nodes with abundant label information, empty-shell virtual nodes without labels may have inappropriate features and lack supervision signals. This assumption can hinder the application of the learned patterns derived from virtual nodes to actual unobserved nodes for kriging.

To address this issue, we propose a Physics-Guided Increment Training Strategy (PGITS), which harnesses domain knowledge about air pollution transport mechanisms to guide the learning process.  Specifically, we utilize real-time wind field data to enrich the attributes of training nodes and develop a dynamic graph generation module. This module incorporates the advection and diffusion processes of airborne particles as physics principles into the graph structure, dynamically adjusting the adjacency matrix to reflect physical interactions between nodes. By incorporating physics knowledge as a link between virtual and real nodes, the features of virtual nodes are closer to actual unobserved nodes. In addition, physical continuity constraints are incorporated into the loss function to guarantee the predictive sequences of the model more closely align with actual environmental changes. Consequently, the learned patterns of virtual nodes can be effectively applied to actual unobserved nodes for accurate kriging. Our contributions are as follows:

(1)We propose a Physics-Guided Increment Training Strategy that integrates physics information into the graph structure, utilizing physics principles as a link between virtual and real nodes.

(2)We design a dynamic graph generation module based on the advection and diffusion processes of airborne particles, dynamically adjusting the adjacency matrix to reflect physical interactions between nodes.

(3)We enhance the predictive accuracy of our deep learning model by integrating specific domain knowledge about air pollution transport mechanisms. Extensive experiments on real-world air quality datasets demonstrate the effectiveness of PGITS.

\section{Related Work}
\subsection{Air Quality Inference}
Air quality inference aims to obtain fine-grained air quality information from sparse observational data. This field has seen extensive research and significant progress over recent decades \cite{han2023machine}. Existing methods are divided into two main categories: physical and data-driven approaches. Physical methods estimate air quality by simulating the propagation and diffusion of pollutants in the atmosphere. Examples of these models include the Gaussian plume models \cite{arystanbekova2004application}, the street canyon models \cite{kim2012urban}, and computational fluid dynamics (CFD) models \cite{huang2008cfd}. These models rely on domain knowledge and often make empirical assumptions that do not align with actual conditions. Data-driven methods estimate air quality by analyzing multisource urban data and capturing the spatio-temporal correlations of air quality distribution. Early statistical methods addressed this issue using K-nearest neighbors (KNN) and Random Forests (RF) \cite{lin2017mining}. However, these methods rely on feature engineering and cannot handle complex non-linear dependencies of pollutant distribution. In recent years, deep learning-based models have demonstrated outstanding performance in air quality inference. ADAIN \cite{cheng2018neural} utilizes data from monitoring stations and urban data closely related to air quality and automatically learns the weights of different monitoring stations for the target area through an attention mechanism. Along this line, MCAM \cite{han2021fine} introduces a multi-channel GNN to model the dynamic and static spatial dependencies between the target and observation areas. Although these models excel in predictive performance, their generalizability may be significantly limited due to the uniqueness and scarcity of external data.

\subsection{Spatio-Temporal Kriging}
Although spatio-temporal kriging is not specifically designed for air quality inference, it has gained significant attention in this field due to its effectiveness in handling spatial and temporal variability. Recent studies categorize spatio-temporal kriging into two distinct frameworks: transductive and inductive kriging. Transductive kriging focuses on estimating values at specific nodes, and all unobserved nodes that require inference must be included in the training phase\cite{zheng2023increase}. In contrast, inductive kriging aims to develop models that generalize to new, unobserved nodes, emphasizing applicability and generalization. Earlier transductive kriging employs matrix factorization \cite{li2020tensor} and tensor completion \cite{lei2022bayesian} to recover the values at the unobserved locations. GE-GAN constructs the graph structure based on observed and unobserved nodes and uses a generative model \cite{goodfellow2020generative} to generate values for the unobserved nodes \cite{xu2020ge}. Recently introduced GRIN integrates message passing mechanism \cite{gilmer2017neural} with GRU \cite{cho2014learning} to capture complex spatio-temporal correlations for kriging \cite{andrea2021filling}. However, these models require retraining due to the limitations of the transductive mechanism when new nodes are added, restricting their application in large-scale dynamic sensor networks. Recent developments in inductive spatio-temporal kriging have shown promising outcomes. KCN \cite{appleby2020kriging} integrates kriging with GNN, demonstrating superior performance. IGNNK \cite{wu2021inductive} utilizes GNN and random subgraphs to master the spatial message passing mechanism, enabling accurate kriging on new graphs without retraining. Additionally, DualSTN \cite{hu2023decoupling} focuses on differentiating long-term and short-term patterns of temporal information, enhancing prediction accuracy through neural network designs. INCREASE \cite{zheng2023increase} adopts a graph representation learning approach that models heterogeneous spatial relations and diverse temporal patterns. Unfortunately, inductive kriging faces the graph gap issue because the training graph is sparser than the inference graph. KITS tries to address this with increment training strategy \cite{xu2023kits}. However, this approach encounters fitting challenges due to the disparity between virtual and real nodes. Especially when executing specific air quality inference tasks, these virtual nodes may learn inappropriate features due to the lack of guidance from domain knowledge.

\section{PRELIMINARIES}
\subsection{Inductive Spatio-temporal Kriging for Air Quality Inference}
\begin{figure}[!t]
\centerline{\includegraphics[width=\columnwidth]{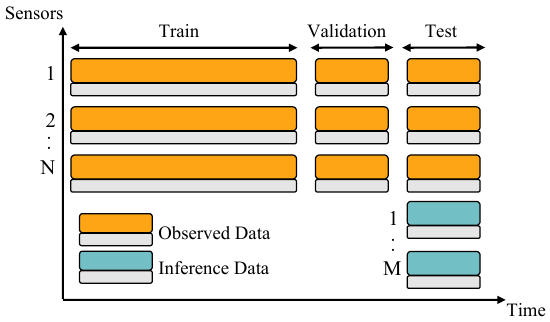}}
\caption{Training and testing strategies of inductive spatio-temporal kriging.}
\label{fig1}
\end{figure}

The goal of air quality inference is to estimate PM\(_{2.5}\) concentrations at \( M\) unmonitored sites using historical concentration data collected at \( N \) monitored stations within the sensor network and other relevant variables. We represent the sensor network as a graph \(G = (\mathrm{V}, \mathrm{E}, \mathbf{W}) \), where \(\mathrm{V} \) is a set of nodes \( |V| = N \), which represents various air quality monitoring stations in the sensor network, \(\mathrm{E} \) is a set of edges between the nodes, \(\mathbf{W} \) is a weighted adjacency matrix, and \(\mathbf{W}_{i, j} \) represents the strengths of the connections between node \( i\) and \( j\) in the graph. Fig. \ref{fig1} displays the training and testing strategies of inductive spatio-temporal kriging. In the training phase, the process includes three steps. Firstly, it constructs a graph over \( N \) observed nodes. Secondly, it randomly masks the values of some nodes, simulating the unobserved nodes. Finally, it learns a function \( h(\cdot) \) to infer data for these masked nodes using the value of unmasked nodes. During testing, it employs \( h(\cdot) \) to estimate the value of  \( M\) unobserved nodes. The learning pattern can be summarized as
\begin{equation}
\mathrm{\hat{X}}_{t-T:t}^{M} = h\left(\mathrm{{X}}_{t-T:t}^{N}, \mathrm{{P}}_{(t-T:t)}^{N}, G\right),
\label{eq:1}
\end{equation}
where \( \mathrm{{X}}_{t-T:t}^{N} \in \mathbb{R}^{N \times T \times 1} \) represents the historical PM\(_{2.5}\) at \( N \) monitored stations, \( \mathrm{{P}}_{(t-T:t)}^{N} \in \mathbb{R}^{N \times T \times 1} \) represents relevant variables, and \( \mathrm{\hat{X}}_{t-T:t}^{M} \in \mathbb{R}^{N \times T \times 1} \) represents the predicted PM\(_{2.5}\) concentrations at \( M \) unmonitored sites.
\begin{figure}[!t]
\centerline{\includegraphics[width=\columnwidth]{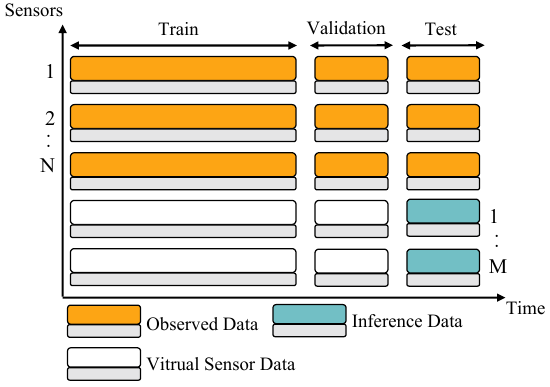}}
\caption{Training and testing strategies of inductive spatio-temporal kriging with increment training strategy.}
\label{fig2}
\end{figure}
\subsection{Inductive Spatio-temporal Kriging with Increment Training Strategy }
In inductive spatio-temporal kriging, the training phase involves \( N \) nodes, while the inference phase incorporates additional \( M \) nodes, making the training graph sparser than the inference graph. The increment training strategy employs virtual nodes to bridge this disparity. Fig. \ref{fig2} illustrates the training and testing settings of inductive spatio-temporal kriging with the increment training strategy. In the training graph, \( M\) empty-shell virtual nodes are added to mimic unobserved nodes requiring inference. Following this configuration, the strategy utilizes the values of observed nodes as labels and employs a semi-supervised approach to estimate the values of virtual nodes, thereby learning the function \( h(\cdot) \). This function is then employed to infer the values of the actual unobserved nodes during the testing phase. The learning pattern of the strategy can be described as follows
\begin{equation}
\mathrm{\hat{X}}_{t-T:t}^{M} = h\left(\mathrm{{X}}_{t-T:t}^{N + M}, \mathrm{{P}}_{(t-T:t)}^{N + M}, G\right),
\label{eq:2}
\end{equation}
where \( \mathrm{{X}}_{t-T:t}^{N + M} \in \mathbb{R}^{(N+M) \times T \times 1} \) represents the historical PM\(_{2.5}\) concentrations at \( N \) observed nodes and \( M \) virtual nodes, \(\mathrm{{P}}_{(t-T:t)}^{N + M} \in \mathbb{R}^{(N+M) \times T \times 2} \) represents relevant variables (e.i., wind speed in U and V directions of each nodes), we represent the sensor network as a graph \(G = (\mathrm{V}, \mathrm{E}, \mathbf{W}) \), where \( \mathrm{V} \) is a set of nodes \( |V| = N + M \), \( \mathrm{E} \in \mathbb{R}^{(N+M) \times(N+M) }\) is a set of edges and \( \mathbf{W} \) is the corresponding weighted adjacency matrix.

\begin{figure*}[!t]
\centerline{\includegraphics[width=\textwidth]{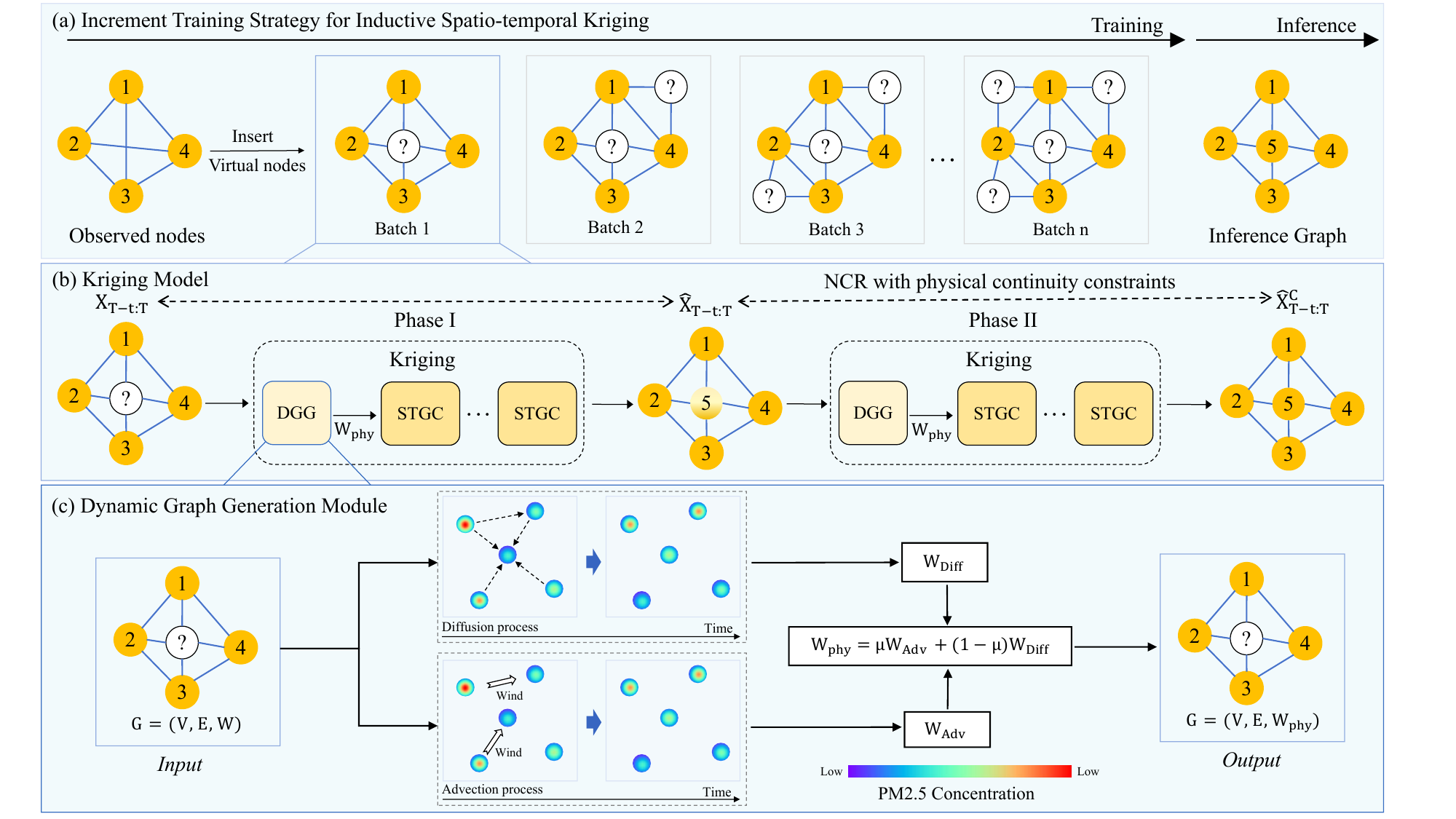}}
\caption{Overview of PGITS. (a) Increment Training Strategy for Inductive Spatio-temporal Kriging (b) Kriging Model with Dynamic Graph Generation (DGG), Spatio-Temporal Graph Convolution (STGC), and Node-Aware Cycle Regulation (NCR) Modules. (c) Dynamic Graph Generation Module Based on the Advection and Diffusion Processes of Airborne Particles. The heatmap represents the PM\(_{2.5}\) concentrations at various nodes, and the arrows indicate the transport of PM\(_{2.5}\) between nodes facilitated by advection and diffusion processes of airborne particles.}
\label{fig3}
\end{figure*}

\section{METHODOLOGY}
\label{sec:guidelines}
This section details the Physics-Guided Increment Training Strategy (PGITS). Fig. \ref{fig3} displays the architecture of the PGITS. According to the increment training strategy shown in Fig. \ref{fig3}(a), we incorporate virtual nodes into the training graph of multiple batches to simulate unobserved nodes that require inference\cite{xu2023kits}. The diversity of the training graph contributes to enhanced generalization of the model across various sensor absence scenarios. Based on these expanded graphs, we employ a two-phase kriging model to estimate the values of all nodes in a semi-supervised manner. As illustrated in Fig. \ref{fig3}(b), we design the kriging model with an encoder-decoder architecture consisting of three components: the Dynamic Graph Generation (DGG) Module, the Spatio-Temporal Graph Convolution (STGC) Module, and the Node-Aware Cycle Regulation (NCR) Module. Specifically, we take the training graph of Batch 1 as an input example to illustrate the operational process of the kriging model. As shown in Fig. \ref{fig3}(c), the DGG incorporates the advection and diffusion processes of airborne particles as physics principles into the graph structure. This integration dynamically adjusts the adjacency matrix to reflect the physical interactions among nodes accurately. We then use the weighted adjacency matrix output from the DGG as input to the STGC and aggregate spatio-temporal features from neighboring nodes to the current node. Ultimately, we incorporate physical continuity constraints into the loss function based on the NCR framework to regulate those unlabeled virtual nodes.

\subsection{Physics Principles of Airborne Particles Movement}
The movement of airborne particles is a key concept in atmospheric science, encompassing the advection and diffusion processes of particulate matter in the atmosphere. These processes collectively determine the paths and extent of atmospheric pollutant propagation. This section details these physical processes and integrates them into the graph structure.

\subsubsection{Advection and diffusion processes of airborne particles}
The movement of airborne particles in space is described using the advection-diffusion equation \cite{kumar2009analytical}, which can be expressed as
\begin{equation}
\frac{\partial \mathrm{X}}{\partial t} + \nabla \cdot (\vec{F} \mathrm{X}) = \mathrm{K} \nabla^2 \mathrm{X},
\label{eq:3}
\end{equation}
where \(\mathrm{X}\) is PM\(_{2.5}\) concentration, \(\vec{F}\) is the flux of particles which describes the transport of PM\(_{2.5}\), and \(\nabla\) is the divergence operator. $\mathrm{K}$ represents the diffusion coefficient.

Based on \eqref{eq:3}, we can describe the two processes of advection and diffusion.  

\textbf{Advection process}: The advection process describes the transport of airborne particles influenced by a flow field, typically driven by a wind field or other large-scale air movement. The flux is represented as a vector field as follows: $\vec{F} =  \vec{v}\, \mathrm{X}$ \cite{hettige2024airphynet}. The advection equation for PM\(_{2.5}\) concentration is given as follows
\begin{equation}
\displaystyle \frac{\partial \mathrm{X}} {\partial t} = - \mathrm{div}\,(\vec{v}\mathrm{X}). 
\label{eq:4}
\end{equation}

\textbf{Diffusion process}: The diffusion process describes the movement of particles along concentration gradients, depicting the transport of airborne particles from areas of high concentration to areas of low concentration. According to \eqref{eq:3}, the diffusion equation for PM\(_{2.5}\) concentration is given as follows
\begin{equation}
\displaystyle \frac{\partial \mathrm{X}}{\partial t} = \mathrm{K}\,\mathrm{div}\,\nabla \mathrm{X}. 
\label{eq:5}
\end{equation}
\subsubsection{Advection and diffusion processes on the graph}
Based on the advection and diffusion processes of airborne particles, we further modeled two physical processes into the graph structure. 

\textbf{Diffusion process on the graph}: The diffusion process on the graph can be represented as the transport of PM\(_{2.5}\) between nodes driven by concentration gradients. The graph Laplacian operator $\Delta$ can be expressed in the gradient dispersion as $\Delta = - \nabla \cdot \nabla$ \cite{bronstein2017geometric}. Therefore, we rewrite the diffusion equation as follows
\begin{equation}
\left( \frac{\partial \mathrm{X}}{\partial t} \right)_{\mathrm{diff}} = -\mathrm{K}\,\Delta \mathrm{X} = -\mathrm{K}\,\mathbf{L}\,\mathrm{X} = -\mathbf{W}_{\mathrm{diff}}\,\mathrm{X},
\label{eq:6}
\end{equation}
where $\mathbf{L}$ is the Laplace matrix of the graph, $\mathbf{W}_{\text{diff}}$ represents the diffusion-based weighted adjacency matrix. In this study, we follow most of the literature for the diffusion coefficient setting and let $\mathrm{K}$ = 0.1 \cite{cussler2009diffusion}.

\textbf{Advection process on the graph}: The advection process on the graph can be modeled by the differences in the wind field data between the nodes, explaining the particle transport due to the influence of the external flow field. In this work, we construct the wind field from the wind speed data of each node in both U and V directions using the formulation proposed by Chapman et al. \cite{chapman2015semi}. The discrete version of the advection equation can be expressed as follows
\begin{equation}
\displaystyle \frac{\partial \mathrm{X}^{i}}{\partial t} = \sum_{\forall j \mid j \to i}  \mathrm{X}^j v_{j \to i} - \sum_{\forall k \mid i \to k} \mathrm{X}^i v_{i \to k} = -[\mathbf{W}_{\mathrm{adv}} \mathrm{X}]^i,
\label{eq:7}
\end{equation}
where \(\sum_{\forall j | j \to i}  \mathrm{X}^j v_{j \to i}\) represents the increase in PM\(_{2.5}\) concentration at node \( i \) in response to the wind field, and \(\sum_{\forall k | i \to k} \mathrm{X}^i v_{i \to k}\) represents the decrease in PM\(_{2.5}\) concentration at node \( i \) in response to the wind field. We consider $\mathbf{W}_{\mathrm{adv}}$ as the advection-based weighted adjacency matrix.

Accordingly, the advection process can be expressed using the weighted adjacency matrix $\mathbf{W}_{\mathrm{adv}}$ based on the flow field as follows
\begin{equation}
\left( \frac{\partial \mathrm{X}}{\partial t} \right)_{\mathrm{adv}} = - \mathbf{W}_{\mathrm{adv}} \mathrm{X},
\label{eq:8} 
\end{equation}
% Where \( W_{adv} \) is the weighted adjacency matrix based on the wind field.
Ultimately, the advection and diffusion processes on the graph can be represented as
\begin{equation}
\frac{\partial \mathrm{X}}{\partial t} = - \mathbf{W}_{\mathrm{diff}} \mathrm{X} - \mathbf{W}_{\mathrm{adv}} \mathrm{X}.
\label{eq:9}
\end{equation}
\subsection{Dynamic Graph Generation (DGG)}
Air quality in real-world scenarios changes dynamically over time and is influenced by environmental variations, particularly the advection and diffusion processes of airborne particles. Therefore, dynamic graph structures are more effective in representing the changing characteristics of sensor networks. As shown in Fig. \ref{fig3}(c), we construct a dynamic graph generation module based on these physical processes, which adaptively fuses advection and diffusion information at different batches of training graphs, reflecting possible physical connections and interactions between nodes through dynamic edge weights.
\subsubsection{The diffusion-based graph}
According to the increment training strategy, we classify the nodes in the training graph into two categories: real nodes and virtual nodes. For real nodes, as the spatially closed locations may share similar data patterns \cite{zheng2023increase}, we use a thresholded Gaussian kernel to connect a node to nearby nodes within a specific radius. The formula is provided as follows
\begin{equation}
\label{eq:10}
    \mathbf{W}^{i,j}_{\mathrm{d}} = \begin{cases}
        \exp\left(-\frac{\mathrm{dist}(i,j)^2}{\gamma}\right), & \mathrm{dist}(i,j) \le \delta \\
        0 , & \mathrm{otherwise},
    \end{cases}
\end{equation}
where $\mathbf{W}^{i,j}_{\mathrm{d}}$ is the weighted adjacency matrix based on the distance of each node, $i$ and $j$ are the indices of two nodes, $dist$ is the geographical distances between nodes, $\gamma$ represents the kernel width (we set it as the standard deviation of distances among nodes), and $\delta$ is the threshold (assigned to 0.1).

We still follow the setup of the KITS \cite{xu2023kits} for virtual nodes. First, we randomly select an observed node. Next, we create a connection between the virtual node and the selected node. Then, we establish connections between the virtual node and each neighboring node of the selected node with a probability of $p \sim \textit{Uniform}[0,1]$. 
Based on the diffusion process on the graph, we can compute the diffusion-based weighted adjacency matrix as follows 
\begin{equation}
    \mathbf{W}_{\mathrm{diff}} = \mathrm{K}\,\mathbf{L} = \mathrm{K}\,(\mathbf{I} - \mathbf{D}^{-\frac{1}{2}} \mathbf{W}_{\mathrm{d}} \mathbf{D}^{-\frac{1}{2}}),
    \label{eq:11}
\end{equation}
where $\mathbf{L}$ is the Laplace matrix of the graph, $\mathbf{W}_{\mathrm{diff}}$ represents the diffusion-based weighted adjacency matrix, and $\mathbf{D}$ is the diagonal degree matrix, diffusion coefficient $\mathrm{K}$ = 0.1.
\subsubsection{The advection-based graph}
Based on the advection process on the graph, we introduce the wind speed for each node in both the \( \mathrm{U} \) and \( \mathrm{V} \) directions, represented by \( \mathrm{{P}}_{(t-T:t)}^{N + M} \in \mathbb{R}^{(N+M) \times T \times 2} \). We introduce wind speed data to the real nodes based on real-world measurements and allocate the wind speed information for the virtual nodes from the areas surrounding the real nodes to which they are connected. This idea stems from the observation that adjacent sensors in the real world are exposed to similar meteorological conditions. We then model the flow field using the wind speed features in both directions. These features are transformed into the high-dimensional space by a Multi-Layer Perceptron (MLP), and the transformations are used to compute edge weights based on their differences. We compute the advection-based weight $\mathbf{W}_{p}^{ij}$ as follows \cite{hettige2024airphynet}
\begin{equation}
\displaystyle p^{i} = {WindField}(P^{i})
\label{eq:12}
\end{equation}
\begin{equation}
\displaystyle p^{j} = {WindField}(P^{j})
\label{eq:13}
\end{equation}
\begin{equation}
\mathbf{W}_{p}^{ij} = p^{i} - p^{j},
\label{eq:14}
\end{equation}
where $\displaystyle P^{i}$ and $\displaystyle P^{j}$ represent the wind-related attributes of the source node $\displaystyle i$ and destination node $\displaystyle j$, respectively, and ${WindField}$ is an MLP used to extract the wind field representation from the corresponding wind data. Based on \eqref{eq:14}, We can compute the advection-based weighted adjacency matrix as follows
\begin{equation}
    \mathbf{W}_{\mathrm{adv}} = \mathbf{I} - \mathbf{D}^{-\frac{1}{2}} \mathbf{W}_p \mathbf{D}^{-\frac{1}{2}},
    \label{eq:15}
\end{equation}
where $\mathbf{W}_{\mathrm{adv}}$ represents the advection-based weighted adjacency matrix, and $\mathbf{D}$ is the diagonal degree matrix.
\subsubsection{Dynamic graph generation process}
We generate multiple batches of training graphs and train them sequentially. Given the wind field variability for each batch, we construct a weighted adjacency matrix based on the advection and diffusion processes to facilitate dynamic graph generation. This dynamism is reflected in the changing edge weights between nodes. The weighted adjacency matrix can be expressed as
\begin{equation}
    \mathbf{W}_{\mathrm{phy}} = \mu \mathbf{W}_{\mathrm{adv}} + (1-\mu) \mathbf{W}_{\mathrm{diff}},
    \label{eq:16}
\end{equation}
where $\mathbf{W}_{\mathrm{phy}}$ represents the weighted adjacency matrix based on the advection and diffusion processes, and $\mu$ is a learnable parameter used to quantify the degree of involvement of advection and diffusion processes.
\subsection{Spatio-Temporal Graph Convolution (STGC)}
The Kriging model is built on the foundation of Spatio-Temporal Graph Convolution (STGC). STGC aggregates spatio-temporal features from neighboring nodes to the current node using graph convolution \cite{andrea2021filling}, which is expressed as follows
\begin{equation}
    \mathrm{Z}_n^{(l+1)} = \textit{FC}(\textit{GC}(\mathrm{Z}_{n-m:n+m}^{(l)}, \mathbf{W}^-)),
    \label{eq:17}
\end{equation} 
where $\mathrm{Z}_{n}^{(l)} \in \mathbb{R}^{(N + M) \times D}$ represents the input feature, with \(n\) denoting different time intervals. To aggregate the features across different time intervals, \(\mathrm{Z}_{n}\) and the features from the previous and following \(m\) time intervals are combined (we assign \(m\) to 1). Here, \( N + M \) is the number of real and virtual nodes, and \( D \) is the feature dimension. The output feature is denoted as $\mathrm{Z}_n^{(l+1)} \in \mathbb{R}^{(N + M) \times D}$, \(l \) and \(l + 1 \) are the layer indices. Additionally, $\textit{FC}(\cdot)$ is the fully connected layer, and $\textit{GC}(\cdot)$ denotes the the inductive graph convolutional layer \cite{andrea2021filling}. 

It is worth noting that \( \mathbf{W}^-\) in \eqref{eq:17} represents the adjacency matrix based on the training graph after removing the self-loops of the nodes, ensuring the features of different nodes can be aggregated based on the training graph. For this part, we replace \( \mathbf{W}^-\) with \( \mathbf{W}_\mathrm{phy}^- \). Therefore, the STGC incorporating physics knowledge is expressed as follows 
\begin{equation}
    \mathbf{Z}_n^{(l+1)} = \textit{FC}(\textit{GC}(\mathbf{Z}_{n-m:n+m}^{(l)}, \mathbf{W}_\mathrm{phy}^-)).
    \label{eq:18}
\end{equation}

According to \eqref{eq:18}, we integrate the advection and diffusion processes of airborne particles as physics knowledge into the existing learning framework.
\subsection{Node-Aware Cycle Regulation (NCR) with Physical Continuity Constraint}
Due to the absence of supervisory signals for the virtual nodes, we utilize the Node-Aware Cycle Regulation (NCR) \cite{xu2023kits} to manage the pseudo-labels used in the learning process, dividing the kriging process into two stages. Initially, this method performs kriging in the first stage (Phase I) to estimate the values of observed and virtual nodes. Subsequently, it switches the roles of observed and virtual nodes with an inverse mask and conducts a second kriging (Phase II), using the output from Phase I as pseudo-labels. Within this framework, we incorporate physical continuity constraint into the loss function to ensure that the model's predictive sequences more closely align with actual environmental changes. Formally, NCR can be written as
\begin{equation}
    \mathrm{\hat{X}}_{t-T:t} = \textit{h}(\mathrm{{X}}_{t-T:t}, \mathbf{W}_\mathrm{phy}^-)
    \label{eq:19}
\end{equation}
\begin{equation}
    \mathrm{X}_{t-T:t}^c = (\mathrm{1} - \mathrm{M}_{t-T:t}) \odot \mathrm{\hat{X}}_{t-T:t}
    \label{eq:20}
\end{equation}
\begin{equation}
    \mathrm{\hat{X}}_{t-T:t}^c = \textit{h}(\mathrm{X}_{t-T:t}^c, \mathbf{W}_\mathrm{phy}^-),
    \label{eq:21}
\end{equation}
where $\mathrm{X}_{t-T:t}$ represents the input data, $\textit{h}(\cdot)$ is the kriging model, $\mathrm{\hat{X}}_{t-T:t}$ is the output from the first stage of the kriging model, $(\mathrm{1} - \mathrm{M}_{t-T:t})$ is the inverse mask, and $\mathrm{\hat{X}}_{t-T:t}^c$ is the final output of the model. This part of the loss function $\mathcal{L}_{sup}$ can be written as
\begin{equation}
\begin{split}
    \mathcal{L}_{\mathrm{sup}} &= \mathrm{MAE}(\mathrm{\hat{X}}_{t-T:t}, \mathrm{X}_{t-T:t}, \mathrm{I}_{\mathrm{obs}}) \\
    &\quad + \lambda \cdot \mathrm{MAE}(\mathrm{\hat{X}}_{t-T:t}^c, \mathrm{\hat{X}}_{t-T:t}, \mathrm{I}_{\mathrm{all}}),
    \label{eq:22}
\end{split}
\end{equation}
where $\mathrm{MAE}(\cdot)$ represents the Mean Absolute Error, $\mathrm{I}_{\mathrm{obs}}$ represents calculating losses on observed nodes, and $\mathrm{I}_{\mathrm{all}}$ mean calculating losses on all nodes.
$\lambda$ is a hyperparameter that determines the weight of pseudo labels (we assign $\lambda$ to 1).

We introduce physical continuity constraints for optimization based on \eqref{eq:22}. The physical loss function $\mathcal{L}_{\mathrm{phy}}$ is defined as 
\begin{equation}
\begin{split}
    \mathcal{L}_{\mathrm{phy}} &= \mathrm{MSE}(\hat{X}_t^c, \hat{X}_{t-1}^c),
    \label{eq:23}
\end{split}
\end{equation}
where $\mathrm{MSE}(\cdot)$ is Mean Square Error. Aiming to guide the model in generating prediction sequences that closely align with actual environmental changes. The final loss function $\mathcal{L}$ can be expressed as

\begin{equation}
\begin{split}
    \mathcal{L} = \mathcal{L}_{\mathrm{sup}} + \beta \cdot \mathcal{L}_{\mathrm{phy}},
    \label{eq:24}
\end{split}
\end{equation}
where $\beta$ is the hyperparameter that controls the importance of the physical continuity constraint.
\section{EXPERIMENTS}
\subsection{DataSets}
We evaluate the performance of the model on real-world air quality dataset AQI-36\cite{zheng2013u, zheng2014urban, zheng2015forecasting}, which consists of hourly data from 36 air quality monitoring stations in Beijing. This dataset covers hourly air quality and meteorological observations from 1 May 2014 to 30 April 2015, including concentrations of major pollutants (PM\(_{2.5}\), PM\(_{10}\), O\(_{3}\), NO\(_{2}\), SO\(_{2}\), and CO), temperature, barometric pressure, humidity, wind speed, and wind direction. This study uses PM\(_{2.5}\) as the target variable, with wind speed as an auxiliary variable.

We follow the methodology outlined in GRIN \cite{andrea2021filling} and ST-MVL \cite{yi2016st}, using data from March, June, September, and December as the test set.

\subsection{Experimental Settings}
\subsubsection{Evaluation Metrics}
The performance of all methods is evaluated using three metrics: Mean Absolute Error (MAE), Mean Absolute Percentage Error (MAPE), and Mean Relative Error (MRE). Their formulas are expressed as follows
\begin{equation}
    \mathrm{MAE} = \frac{1}{|\Omega|}\sum_{i \in \Omega}|\mathrm{Y}_i-\mathrm{\hat{Y}}_i|
\end{equation}
\begin{equation}
    \mathrm{MAPE} = \frac{1}{|\Omega|}\sum_{i \in \Omega}\frac{|\mathrm{Y}_i-\mathrm{\hat{Y}}_i|}{|\mathrm{Y}_i|}
\end{equation}
\begin{equation}
    \mathrm{MRE} = \frac{\sum_{i \in \Omega}|\mathrm{Y}_i-\mathrm{\hat{Y}}_i|}{\sum_{i \in \Omega}|\mathrm{Y}_i|},
\end{equation}
where $\Omega$ represents the index set of unobserved nodes used for evaluation, $\mathrm{Y}$ denotes the ground truth data, and $\mathrm{\hat{Y}}$ indicates the predictions produced by the kriging models.
\subsubsection{Experimental Settings}
Our model is implemented by PyTorch 1.8.1 with NVIDIA GeForce RTX 3080ti GPU. We trained our model with the Adam optimizer, setting the batch size to 32 and the initial learning rate to 0.0002. Following KITS \cite{xu2023kits} (one of the baseline methods), we set the missing rate $\alpha$ = 0.5 (the proportion of unobserved nodes to all nodes in the inference graph) in all datasets. The window size for the time series is set to 24 and the feature dimension is 64. We employ the early stopping mechanism to mitigate overfitting to preserve the best-performing models based on validation set performance.
\subsection{Baselines}
\begin{itemize}
\item {KNN:}\ K-Nearest Neighbors, a simple baseline averaging the nearest geographical neighbors. 

\item {KCN \cite{appleby2020kriging}:}\ Kriging Convolutional Network is a spatial interpolation method based on graph neural networks. Spatial interpolation is achieved by aggregating information from the K nearest neighbors of the target node using the graph neural network.

\item {IGNNK \cite{wu2021inductive}:}\ Inductive Graph Neural Network Kriging utilizes GNN and random subgraphs to master the spatial message passing mechanism. The value of the current node is interpolated using the spatio-temporal information from nearby nodes.

\item {DualSTN \cite{hu2023decoupling}:}\ The Dual Joint SpatioTemporal Network is similar to IGNNK, but it specifically focuses on differentiating long-term and short-term patterns of temporal information.

\item {INCREASE \cite{zheng2023increase}:}\ Inductive Graph Representation Learning for Spatio-Temporal Kriging. It adopts a graph representation learning approach that models heterogeneous spatial relations and diverse temporal patterns.

\item {KITS \cite{xu2023kits}:}\ Inductive Spatio-Temporal Kriging With Increment Training Strategy. KITS is the latest inductive spatio-temporal kriging method, utilizing virtual nodes to simulate the unobserved nodes during the training phase.
\end{itemize}
\subsection{Main Experimental Results}
\subsubsection{Overall Comparison}
Table \ref{tab:model_aqi_comparison} compares our proposed PGITS with six baseline methods on the air quality kriging tasks. Our model outperforms all baseline models in three predictive metrics, including MAE, MAPE, and MRE reported by Xu et al. \cite{xu2023kits}. For example, the MAPE decreased by 5.13$\%$. We attribute this to the integration of domain knowledge, which reduces the discrepancy between virtual and real nodes, and the integration of physical information effectively mitigates fitting issues caused by node differences. In practical applications, the dynamic graph generation module significantly enhances the utility of the model by more accurately capturing the dynamic features of sensor networks. Furthermore, the introduction of physical continuity constraints reinforces the consistency between the model predictions and real environmental variations. These results substantiate the effectiveness and potential of the model for air quality inference in complex environments and demonstrate its promising applicability.

\begin{table}[h]
  \centering
  \vspace{-10pt}
  \caption{Performance comparison with inductive kriging baselines on the AQI-36 dataset. }
  \label{tab:model_aqi_comparison}
  \scriptsize 
  \renewcommand{\arraystretch}{0.68} 
  \resizebox{0.48\textwidth}{!}{
    \begin{tabular}{r|ccc}
    \midrule
    \multirow{2}{*}{Model} & \multicolumn{3}{c}{AQI-36} \\
    \cmidrule{2-4}
          & MAE   & MAPE  & MRE  \\
    \midrule 
    KNN   & 18.35 & 0.50 & 0.24 \\
    KCN   & 20.64 & 0.62 & 0.29 \\
    IGNNK & 23.35 & 0.78 & 0.31 \\
    DualSTN & 22.77 & 0.90 & 0.32 \\
    INCREASE & 22.90 & 1.07 & 0.32 \\
    KITS  & 16.59 & 0.39 & 0.24 \\
   \textbf{PGITS (ours)}  & \textbf{16.36} & \textbf{0.37} & \textbf{0.23} \\
    \midrule
    Improvements & 1.39\% & 5.13\% & 4.17\% \\
    \midrule
    \end{tabular}
  }
  %\vspace{-10pt}
\end{table}

\section{Conclusion}
In this paper, we propose a Physics-Guided Increment Training Strategy to address the disparity between virtual and real nodes in the increment training strategy. This strategy utilizes physics principles as a bridge between virtual and real nodes, ensuring that the features of virtual nodes and their pseudo labels are closer to actual nodes. Specifically, we design a dynamic graph generation module to incorporate the advection and diffusion processes of airborne particles as physics knowledge into the graph structure, dynamically adjusting the adjacency matrix to reflect physical interactions between nodes. Consequently, the learned patterns derived from virtual nodes can be applied to actual unobserved nodes for effective kriging. We conduct extensive experiments on real-world air quality dataset, demonstrating the effectiveness of PGITS. In future work, we will consider integrating more complex spatio-temporal physics principles into existing deep learning frameworks.

\bibliographystyle{IEEEtran}
\bibliography{IEEEabrv,Ref.bib}
\end{document}